\documentclass[letterpaper, 10 pt, conference]{ieeeconf}  

\IEEEoverridecommandlockouts                              

\usepackage{graphics} 
\usepackage{epsfig} 
\usepackage{mathptmx} 
\usepackage{times} 
\usepackage{amsmath} 
\usepackage{amssymb}  
\usepackage{lscape}
\usepackage{booktabs}
\usepackage{stfloats}
\usepackage{multirow}
\usepackage{graphicx}
\usepackage{utfsym}
\usepackage{url}

\title{\LARGE \bf
	Similarity Distance-Based Label Assignment for Tiny Object Detection
}

\author{Shuohao Shi$^{1, \dag}$, Qiang Fang$^{1, \dag, *}$, Xin Xu$^{1}$, Tong Zhao$^{1}$
	\thanks{$^{1}$National University of Defense Technology, Changsha, Hunan, China
		{\tt\small e867cda2b@126.com, qiangfang@nudt.edu.cn, xinxu@nudt.edu.cn, zhaotong@nudt.edu.cn}}%
	\thanks{$^{\dag}$These authors contributed equally and should be co-first authors}
	\thanks{$^{*}$corresponding author}
}

\begin{document}

	\maketitle
	\thispagestyle{empty}
	\pagestyle{empty}

	\begin{abstract}
		
		Tiny object detection is becoming one of the most challenging tasks in computer vision because of the limited object size and lack of information.
		The label assignment strategy is a key factor affecting the accuracy of object detection.
		Although there are some effective label assignment strategies for tiny objects, most of them focus on reducing the sensitivity to the bounding boxes to increase the number of positive samples and have some fixed hyperparameters need to set.
		However, more positive samples may not necessarily lead to better detection results, in fact, excessive positive samples may lead to more false positives.
		In this paper, we introduce a simple but effective strategy named the Similarity Distance (SimD) to evaluate the similarity between bounding boxes.
		This proposed strategy not only considers both location and shape similarity but also learns  hyperparameters adaptively, ensuring that it can adapt to different datasets and various object sizes in a dataset.
		Our approach can be simply applied in common anchor-based detectors in place of the IoU for label assignment and Non Maximum Suppression (NMS).
		Extensive experiments on four mainstream tiny object 
		detection datasets demonstrate superior performance of our method, especially, 1.8 AP points and 4.1 AP points of very tiny higher than the state-of-the-art competitors on AI-TOD.
		Code is available at: \url{https://github.com/cszzshi/SimD}.
		
	\end{abstract}

	\section{INTRODUCTION}
	
	With the popularization of drone technology and autonomous driving, applications of object detection are becoming increasingly widespread in daily life.
	General object detectors have achieved significant progress in both accuracy and detection speed.
	For example, the latest version of the YOLO series, YOLOv8, achieves a mean average precision (mAP) of 53.9 percent on the COCO detection dataset and takes only 3.53 ms to detect objects in an image when implemented on the NVIDIA A100 GPU using TensorRT.
	Nevertheless, despite this significant progress in general object detectors, when they are directly applied for tiny object detection tasks, their accuracy sharply decreases.
	
	In a recent survey of small object detection, Cheng et al. \cite{ref1} proposed dividing small objects into three categories (extremely, relatively and generally small) in accordance with their mean area.
	Two major challenges faced in tiny object detection are information loss and the lack of positive samples.
	There are many possible approaches to improve the accuracy of tiny object detection, such as feature fusion, data augmentation, and superresolution.
	
	\begin{figure}[t]
		\centering
		\includegraphics[,width=8.2cm]{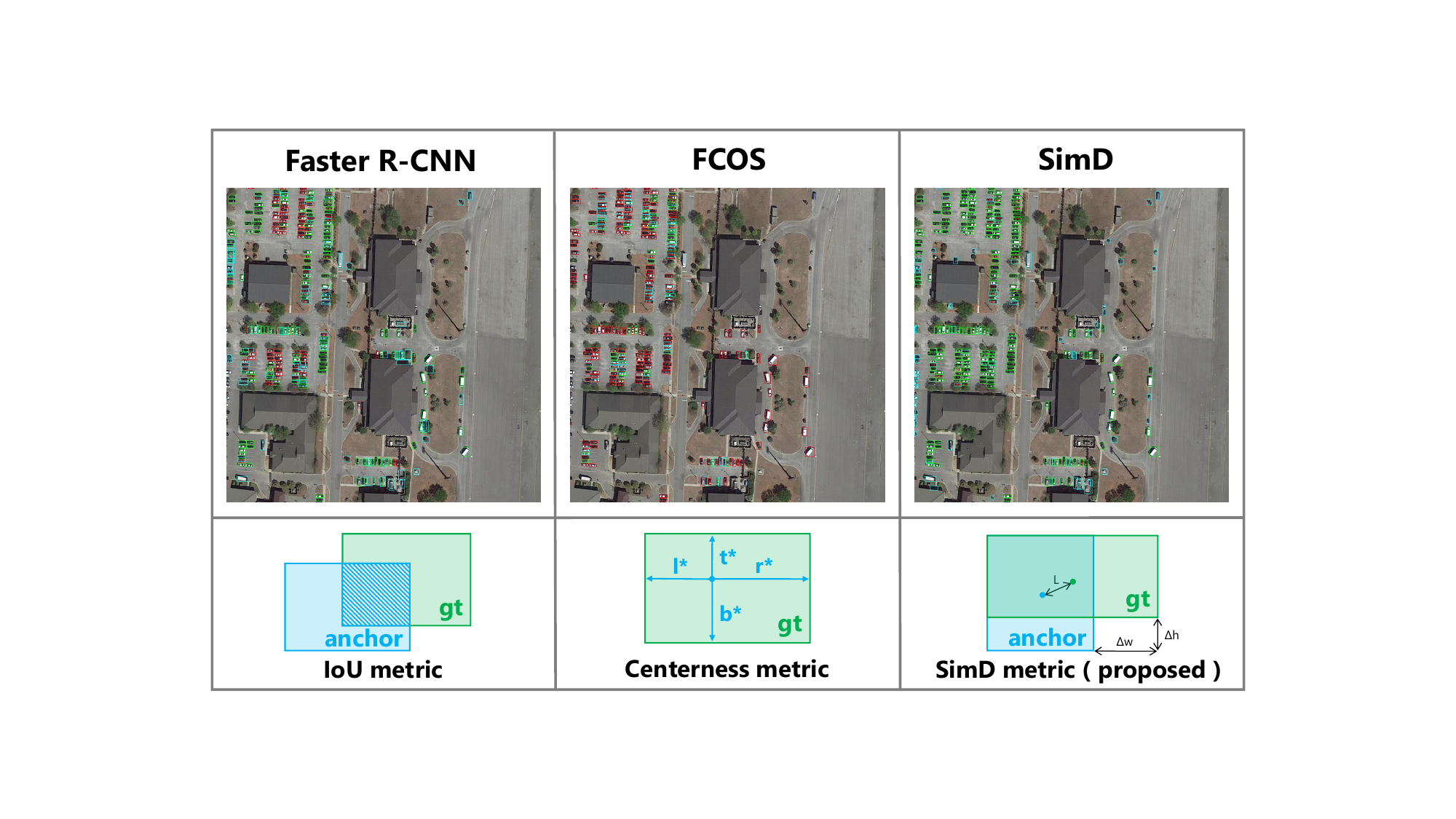}
		\caption{
			Comparison between traditional label assignment metrics and our SimD metric.
			The first row shows typical detection results achieved with these methods, and the second row presents diagrammatic sketches of these metrics.
			The $\Delta w$ and $\Delta h$ in SimD respectively represent the difference of width and height between anchor and ground truth.
			The green, blue and red boxes respectively denote true positive (TP), false positive (FP) and false negative (FN) predictions.
		}
		\label{compare_simd_fr_fcos}
	\end{figure}
	
	Because sufficiently numerous and high-quality positive samples are crucial for object detection, the label assignment strategy is a core factor affecting the final results.
	The smaller the bounding boxes are, the higher the sensitivity of the IoU metric \cite{ref2}, this is the main reason why it is not possible to label as many positive samples as tiny objects as can be labeled as general objects.
	A simple comparison between traditional anchor-based and anchor-free metrics and our SimD metric is shown in Fig. \ref{compare_simd_fr_fcos}.
	
	The current research on tiny object label assignment strategies mainly focuses on reducing the sensitivity to the bounding box size.
	From this perspective, Xu et al. \cite{ref2} proposed using the Dot Distance (DotD) as the assignment metric in place of the IoU.
	Later, NWD \cite{ref3} and RFLA \cite{ref4} were proposed as attempts to model the ground truth and anchor as Gaussian distributions and then use the distance between these two Gaussian distributions to evaluate the two bounding boxes.
	In fact, these methods have enabled considerable progress in label assignment, but there are also some problems that they may not consider.
	
	First, most of these methods focus on reducing the sensitivity to the bounding box size, thereby increasing the number of positive samples.
	However, as we know, excessive positive samples may have a detrimental impact on an object detector, leading to many false positives.
	
	\begin{figure*}[t]
		\centering
		\includegraphics[,width=17.3cm]{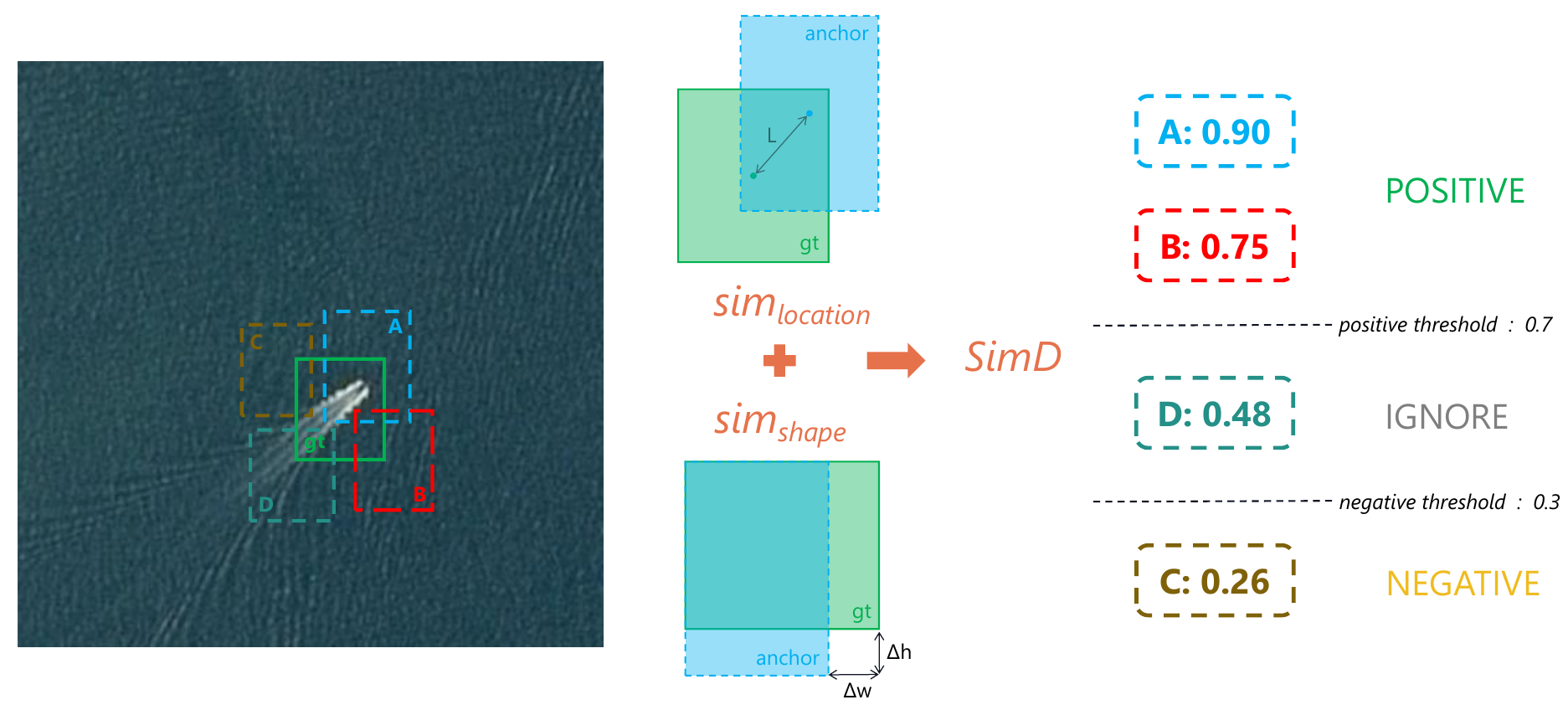}
		\caption{
			The processing flow of the SimD-based label assignment strategy.
			We first obtain the coordinates of the ground truth and anchors and then calculate the Similarity Distance (SimD) between the ground truth and each anchor. Subsequently, we follow the traditional label assignment strategy to obtain positive and negative samples in accordance with corresponding thresholds.
			For a ground truth that does not have any associated positive sample based on this strategy, we assign the anchor with the maximum SimD value as a positive sample, as long as this SimD value is larger than a minimum positive threshold.
		}
		\label{SimD_stream}
	\end{figure*}
	
	Second, the essence of these evaluation metrics is to measure the similarity between bounding boxes. For anchor-based methods, the similarity between the ground truth and the anchor is considered.
	This similarity includes two aspects: shape and location. However, some methods only consider the locations of the bounding boxes, the others consider both shape and location, but they also have a hyperparameter that needs to be chosen.
	
	Finally, although the object sizes in a tiny object detection dataset tend to be fairly similar, there are still differences in the scales of different objects.
	For example, the sizes of the objects in the AI-TOD dataset range from 2 to 64 pixels. The discrepancy is more pronounced in the VisDrone2019 dataset, which contains both small and general-sized objects. 
	In fact, the smaller the size of object, the more difficult it is to obtain positive samples.
	Unfortunately, most of the existing methods may pay less attention to this problem.
	
	In this paper, to solve these problems, we introduce a new evaluation metric to take the place of the traditional IoU, the processing flow of our method is shown in Fig. \ref{SimD_stream}.
	The main contributions of this paper include the following:
	
	\textbullet \; We propose a simple but effective strategy named Similarity Distance (SimD) to evaluate the relationship between two bounding boxes.
	It not only considers both location and shape similarity but also can effectively adapt to different datasets and different object sizes in a dataset without the need to set any hyperparameter.
	
	\textbullet \; Extensive experiments prove the validity of our method.
	We use several general object detectors and simply replace the IoU-based assignment module with the proposed method based on our SimD metric, in this way, we achieve the state-of-the-art performance on four mainstream tiny object detection datasets.
	
	\section{RELATED WORK}
	
	In recent years, applications of object detection have become increasingly widespread in various industries.
	This technology offers considerable convenience.
	For example, rescue operations can be quickly carried out by identifying ground objects in remote sensing images.
	With the development of deep learning technology, especially the introduction of ResNet \cite{ref5}, the accuracy and speed of detection have significantly increased.
	
	General object detectors can be divided into two categories: one-stage and two-stage detectors.
	A two-stage detector first generates a list of proposal regions and then determines the position and category of the object.
	Such algorithms include R-CNN \cite{ref6}, Fast R-CNN \cite{ref7} and Faster R-CNN \cite{ref8}.
	The structure of one-stage detectors is simpler.
	They can directly output the coordinates and category of an object from the input image.
	Some classic one-stage detectors include YOLO \cite{ref9} and SSD \cite{ref10}.
	
	\subsection{Tiny Object Detection}
	
	Despite the significant progress in object detection achieved with deep learning technology, the detection accuracy will sharply decrease when the objects to be detected are tiny.
	Small objects are usually defined as objects with sizes less than a certain threshold value.
	For example, in Microsoft COCO \cite{ref11}, if the area of an object is less than or equal to 1024, it is considered a small object.
	In many cases, however, the objects of interest are in fact much smaller than the above definition.
	For example, in the AI-TOD dataset, the average edge length of an object is only 12.8 pixels, far smaller than in other datasets.
	
	As stated in a previous paper \cite{ref1}, due to the extremely small size of the objects of interest, there are three main challenges in tiny object detection.
	First, most object detectors use downsampling for feature extraction, which will result in a large amount of information loss for tiny objects.
	Second, due to the limited amount of valid information they contain, small objects are easily disturbed by noise.
	Finally, the smaller an object is, the more sensitive it is to changes in the bounding box \cite{ref2}.
	Consequently, if we use traditional label assignment metrics, such as the IoU, GIoU \cite{ref12}, DIoU \cite{ref13}, and CIoU \cite{ref13}, for object detection, the number of positive samples obtained for tiny objects will be very small.
	
	Many methods and have been proposed to improve the accuracy and efficiency of tiny object detection.
	For example, from the perspective of data augmentation, Kisantal et al. \cite{ref14} proposed increasing the number of training samples by copying tiny objects, randomly transforming the copies and then pasting the results into new positions in an image.
	
	\subsection{Label Assignment Strategies}
	
	The label assignment strategy plays a significant role in object detection.
	Depending on whether each label is either strictly negative or strictly positive, such strategies can be divided into hard label assignment strategies and soft label assignment strategies.
	In a soft label assignment strategy, different weights are set for different samples based on the calculation results, examples include GFL \cite{ref15}, VFL \cite{ref16}, TOOD \cite{ref17} and DW \cite{ref18}.
	Hard label assignment strategies can be further divided into static and dynamic strategies depending on whether the thresholds for designating positive and negative samples are fixed.
	Static label assignment strategies include those based on the IoU and DotD \cite{ref2} metrics as well as RFLA \cite{ref4}.
	Examples of dynamic label assignment strategies include ATSS \cite{ref19}, PAA \cite{ref20}, OTA \cite{ref21}, and DSLA \cite{ref22}.
	From another perspective, label assignment strategies can be divided into prediction-based and prediction-free strategies.
	A prediction-based method assigns a sample a positive/negative label based on the relationship between the ground-truth and predicted bounding boxes, while a prediction-free method assigns labels based only on the anchors or other existing information.
	
	\subsection{Label Assignment Strategies for Tiny Objects}
	
	\begin{table*}[t]
		\centering
		\caption
		{
			Comparison between existing label assignment methods for tiny objects and our method
		}
		\label{formula}
		\begin{tabular}{c|c|c|c|c}
			\toprule
			Method                    & Formula                                                                & Insensitive           & Comprehensive         & Adaptive \\ \midrule
			\multicolumn{1}{c|}{DotD} & \multicolumn{1}{c|}{\begin{tabular}[c]{@{}c@{}}$DotD=e^{-\frac{D}{S} } $, $D=\sqrt{\left ( x_{g} -x_{a}  \right )^{2} +\left ( y_{g} - y_{a}   \right ) ^{2}  } $, $S=\sqrt{\frac{ {\textstyle \sum_{i=1}^{M} {\textstyle \sum_{j=1}^{N_{i} }w_{ij}\times h_{ij}  } } }{ {\textstyle \sum_{i=1}^{M}}N_{i}  } } $\end{tabular}} & \multicolumn{1}{c|}{\checkmark} & \multicolumn{1}{c|}{\usym{2613}} & \usym{2613}           \\ \midrule
			\multicolumn{1}{c|}{NWD}  & \multicolumn{1}{c|}{\begin{tabular}[c]{@{}c@{}} $NWD=e^{-\frac{W}{C} }$, $W=\sqrt{\left ( x_{g} -x_{a}  \right )^{2} + \left ( y_{g} -y_{a}  \right )^{2}+\left ( \left ( w_{g} -w_{a}  \right )^{2} + \left ( h_{g} -h_{a}  \right )^{2} \right )\times 0.25 } $ \end{tabular}} & \multicolumn{1}{c|}{\checkmark} & \multicolumn{1}{c|}{\checkmark} &  \usym{2613}        \\ \midrule
			\multicolumn{1}{c|}{RFLA} & \multicolumn{1}{c|}{\begin{tabular}[c]{@{}c@{}}$RFD=\frac{1}{1+RFDC} $, $RFDC=\frac{0.5 \beta \times w_{a}^{2} }{w_{g}^{2} } +\frac{0.5 \beta \times h_{a}^{2}  }{h_{g}^{2} }+\frac{2\left ( x_{a}-x_{g}   \right )^{2}  }{w_{g}^{2} } +\frac{2\left ( y_{a}-y_{g}   \right )^{2}  }{h_{g}^{2} } +\ln_{}{\frac{w_{g}}{\beta \times w_{a}}} +\ln_{}{\frac{h_{g}}{\beta \times h_{a}}} -1 $\\ \end{tabular}}     & \multicolumn{1}{c|}{\checkmark} & \multicolumn{1}{c|}{\checkmark} &  \usym{2613}        \\ \midrule
			\multicolumn{1}{c|}{SimD} & \multicolumn{1}{c|}{\begin{tabular}[c]{@{}c@{}}$	SimD=e^{-\left ( sim_{location} + sim_{shape}   \right ) }$\end{tabular}} & \multicolumn{1}{c|}{\checkmark} & \multicolumn{1}{c|}{\checkmark} & \checkmark         \\ \bottomrule
		\end{tabular}
	\end{table*}
	
	Although there have been many studies on label assignment strategies for object detection, most such strategies are designed for conventional datasets, with few specifically designed for tiny objects.
	When these traditional label assignment strategies are directly used for tiny object detection, they suffer a significant decrease in accuracy.
	To date, the label assignment strategies and metrics designed specifically for tiny objects mainly include $S^3$FD \cite{ref22_1}, DotD \cite{ref2}, NWD-RKA \cite{ref23} and RFLA \cite{ref4}.
	
	In $S^3$FD, the threshold value is first reduced (from 0.5 to 0.35) to obtain more positive samples for ground truth and then is further lowered to 0.1 to obtain positive samples for those ground truths that were not addressed with the first threshold reduction.
	However, $S^3$FD also uses the traditional IoU metric to compute the similarity between the ground truth and anchor.
	To overcome the weakness of the IoU metric, the novel DotD formula was introduced to reduce the sensitivity to the bounding box size.
	Based on this metric, more positive samples can be obtained for the ground truth.
	In NWD-RKA, a normalized Wasserstein distance is introduced as a replacement for the IoU, and a ranking-based strategy is used to assign the top-k samples as positive.
	RFLA explores the relationship between the ground truth and anchor from the perspective of the receptive field, on this basis, the ground truth and anchor are modeled as Gaussian distributions.
	Then, the distance between these two Gaussian distributions is calculated based on the Kullback--Leibler divergence (KLD), which is used in place of the IoU metric. 
	
	\section{Method}
	
	\subsection{Similarity Distance Between Bounding Boxes}
	
	One of the most important steps in label assignment is to calculate a value that reflects the similarity between different bounding boxes.
	Specifically, for an anchor-based label assignment strategy, the similarity between the anchors and ground truths must be quantified before assigning labels.
	
	Common label assignment metrics, such as the IoU, GIoU \cite{ref12}, DIoU \cite{ref13} and CIoU \cite{ref13}, are usually based on the overlap between the anchor and the ground truth. These metrics have the serious problem that if the overlap is zero, which is often the case for tiny objects, these metrics may become invalid.
	Some more suitable methods use distance-based evaluation metrics or even use Gaussian distributions to model the ground truth and anchor, such as the DotD \cite{ref2}, NWD \cite{ref3} and RFLA \cite{ref4}.
	We present a simple comparison between the existing metrics and our SimD metric from three perspectives in Table \ref{formula}.
	For example, DotD only considers location similarity and may not adapt different object sizes in a dataset, so it is not comprehensive or adaptive.
	NWD and RFLA are not adaptive because they respectively have a hyperparameter $C$ and $\beta$ need to set.
	Following the existing approaches, we consider proposing an adaptive method without any hyperparameters.
	
	In this paper, we introduce a novel metric named Similarity Distance (SimD) to better reflect the similarity between different bounding boxes.
	The Similarity Distance is defined as follows:
	\begin{equation}
		SimD=e^{-\left ( sim_{location} + sim_{shape}   \right ) }   \label{eqsim}
	\end{equation}
	\begin{equation}
		sim_{location}=\sqrt{\left ( \cfrac{x_{g}-x_{a}} {\frac{1}{m}\times \left ( w_{g}+w_{a} \right )   }  \right )^{2}  +\left ( \cfrac{y_{g}-y_{a}} {\frac{1}{n}\times \left ( h_{g}+h_{a} \right )   }  \right )^{2}  } \label{eqsimloc}
	\end{equation}
	\begin{equation}
		sim_{shape}=\sqrt{\left ( \cfrac{w_{g}-w_{a}} {\frac{1}{m}\times \left ( w_{g}+w_{a} \right )   }  \right )^{2}  +\left ( \cfrac{h_{g}-h_{a}} {\frac{1}{n}\times \left ( h_{g}+h_{a} \right )   }  \right )^{2}  } \label{eqsimshape}
	\end{equation}
	where $m$ and $n$ are shown below:
	\begin{equation}
		m=\cfrac{ {\sum_{i=1}^{M}}  {\sum_{j=1}^{N_{i} } {\sum_{k=1}^{Q_{i}}\cfrac{\left | x_{ij} -x_{ik} \right |  }{w_{ij}+w_{ik}  } } }  }{ {\sum_{i=1}^{M}}N_{i} \times Q_{i}  }  \label{eqm} 
	\end{equation}
	\begin{equation}
		n=\cfrac{ {\sum_{i=1}^{M}}  {\sum_{j=1}^{N_{i} } {\sum_{k=1}^{Q_{i}}\cfrac{\left | y_{ij} -y_{ik} \right |  }{h_{ij}+h_{ik}  } } }  }{ {\sum_{i=1}^{M}}N_{i} \times Q_{i}  } \label{eqn} 
	\end{equation}
	
	The SimD contains two parts, location similarity, $sim_{location}$ and shape similarity, $sim_{shape}$. 
	As shown in (\ref{eqsimloc}), ($x_g$, $y_g$) and ($x_a$, $y_a$) respectively represent the center coordinate of ground truth and anchor, $w_g$, $w_a$, $h_g$, $h_a$ represent the width and height of ground truth and anchor, the main part of the formula is the distance between the center point of ground truth and anchor, similar to the DotD formula \cite{ref2}.
	The difference is that we use the widths and heights of the two bounding boxes multiplied by corresponding parameters to eliminate the differences between bounding boxes of different sizes.
	This process is similar to the idea of normalization.
	This is also why our metric can easily adapt to different object sizes in a dataset.
	The shape similarity in (\ref{eqsimshape}) is similar to (\ref{eqsimloc}).
	
	The definitions of the two normalization parameters are shown in (\ref{eqm}) and (\ref{eqn}).
	Because they are quite similar, we take the parameter $m$ in (\ref{eqm}) as an example for further discussion. $m$ is the average ratio of the distance in the $x$-direction to the sum of the two widths for all ground truths and anchors in each image in the whole train set.
	$M$ represents the number of images in the train set,  $N_{i}$ and $Q_{i}$ represent the number of ground truths and anchors, respectively, in the $i$-th image.
	$x_{ij}$ and $x_{ik}$ respectively represent the $x$-coordinate of the center point of $j$-th ground truth and $k$-th anchor in $i$-th image.
	$w_{ij}$ and $w_{ik}$ respectively represent the width of $j$-th ground truth and $k$-th anchor in $i$-th image.
	Because the two normalization parameters are computed based on the train set, our metric can also automatically adapt to different datasets.
	
	To facilitate label assignment, we use an exponential function to scale the value of SimD to the range between zero and one.
	If the two bounding boxes are the same, the value below the root sign will be zero, so SimD will be equal to one.
	If the two bounding boxes greatly differ, this value will be very large, so SimD will approach zero.
	
	\subsection{Similarity Distance-based Detectors}
	
	The novel SimD metric defined in (\ref{eqsim}) can well reflect the relationship between two bounding boxes and is easy to compute.
	Therefore, it can be used in place of the IoU in scenarios where the similarity between two bounding boxes needs to be calculated.
	
	\textbf{SimD-based label assignment.}
	In traditional object detectors, such as Faster R-CNN \cite{ref8}, Cascade R-CNN \cite{ref24} and DetectoRS \cite{ref25}, the label assignment strategy for the RPN and R-CNN models is MaxIoUAssigner.
	MaxIoUAssigner considers three threshold values: a positive threshold, a negative threshold and a minimum positive threshold.
	Anchors for which the IoU with respect to the ground truth is higher than the positive threshold are positive samples, those for which the IoU is lower than the negative threshold are negative samples, and those for which the IoU is between the positive threshold and the negative threshold are ignored.
	For tiny object detection, Xu et al. introduced the RKA \cite{ref23} and HLA \cite{ref4} label assignment strategies, which do not use fixed thresholds to divide positive and negative samples.
	In RKA, the top-k anchors associated with a ground truth are simply selected as positive samples, this strategy can increase the number of positive samples because the assignment of positive labels is not limited by a positive threshold.
	However, introducing too many low-quality positive samples may cause the detection accuracy to degrade.
	
	In this paper, we follow the traditional MaxIoUAssigner strategy and simply use SimD in place of the IoU. The positive threshold, negative threshold, and minimum positive threshold are set to 0.7, 0.3, and 0.3, respectively.
	Our label assignment strategy is named MaxSimDAssigner.
	
	\textbf{SimD-based NMS.}
	Non Maximum Suppression (NMS) is one of the most important components of postprocessing.
	Its purpose is to eliminate predicted bounding boxes that are repeatedly detected by preserving only the best detection result.
	In the traditional NMS procedure, first, the IoUs between the bounding box with the highest score and all other bounding boxes are computed.
	Then, bounding boxes with an IoU higher than a certain threshold will be eliminated.
	Considering the advantages of SimD, we can simply use it as the metric for NMS in place of the traditional IoU metric.
	
	\section{Experiments}
	
	To verify the reliability of our proposed method, we designed a series of experiments involving the application of traditional object detectors to several open-source tiny object detection datasets.
	
	\subsection{Datasets}
	
	There are two main types of tiny object detection datasets: one contains only small objects, such as AI-TOD \cite{ref26}, AI-TODv2 \cite{ref23}, and SODA-D \cite{ref1}, the other contains both small and medium-sized objects, such as VisDrone2019 \cite{ref27} and TinyPerson \cite{ref28}.
	
	\textbf{AI-TOD.}
	AI-TOD (Tiny Object Detection in Aerial Images) is an aerial remote sensing small object detection dataset collected to address the shortage of available datasets for aerial image object detection tasks.
	It contains 28,036 images and 700,621 object instances divided into eight categories with accurate annotations.
	Due to the extremely small size of its object instances (the mean size is only 12.8 pixels), it can be effectively used to test the capabilities of tiny object detectors.
	
	\textbf{SODA-D.}
	The SODA (Small Object Detection dAtasets) series includes two datasets: SODA-A and SODA-D.
	SODA-D was collected from MVD \cite{ref29}, which consists of images captured from streets, highways and other similar scenes.
	There are 25,834 extremely small objects (with areas ranging from 0 to 144) in SODA-D \cite{ref1}, making it an excellent benchmark for tiny object detection tasks.
	
	\textbf{VisDrone2019.}
	VisDrone2019 is a dataset from the VisDrone Object Detection in Images Challenge.
	For this competition, 10,209 static images were captured by an unmanned drone in different locations at different heights and angles.
	VisDrone2019 is also an excellent dataset for evaluating tiny object detectors because it contains not only extremely small objects but also normally sized objects.
	
	\begin{table*}[t]
		\centering
		\caption
		{
			Main results on AI-TOD.
			All models are trained on the trainval set and tested on the test set.
			AP$_{vt}$, AP$_{t}$, AP$_{s}$ and AP$_{m}$ respectively represent the average precision for very tiny (2 to 8 pixels), tiny (8 to 16 pixels), small (16 to 32 pixels) and medium (32 to 64 pixels) objects
		}
		\label{AI-TOD}
		\begin{tabular}{l|c|ccc|cccc}
			\toprule
			Method                & Backbone  & AP   & AP$_{0.5}$ & AP$_{0.75}$ & AP$_{vt}$ & AP$_{t}$  & AP$_{s}$  & AP$_{m}$  \\ \midrule
			TridentNet \cite{ref32}            & ResNet-50 & 7.5  & 20.9  & 3.6    & 1.0  & 5.8  & 12.6 & 14.0 \\
			Faster R-CNN \cite{ref8}          & ResNet-50 & 11.1 & 26.3  & 7.6    & 0.0  & 7.2  & 23.3 & 33.6 \\
			Cascade R-CNN \cite{ref24}         & ResNet-50 & 13.8 & 30.8  & 10.5   & 0.0  & 10.6 & 25.5 & 36.6 \\
			DetectoRS \cite{ref25}             & ResNet-50 & 14.8 & 32.8  & 11.4   & 0.0  & 10.8 & 18.3 & 38.0 \\
			DotD \cite{ref2}                     & ResNet-50  & 16.1 & 39.2  & 10.6   & 8.3  & 17.6 & 18.1 & 22.1 \\
			DetectoRS w/ NWD \cite{ref3}      & ResNet-50 & 20.8 & 49.3  & 14.3   & 6.4  & 19.7 & 29.6 & \textbf{38.3} \\ 
			DetectoRS w/ RFLA \cite{ref4}     & ResNet-50 & 24.8 & 55.2  & 18.5   & 9.3  & 24.8 & 30.3 & 38.2 \\ \midrule
			SSD-512 \cite{ref10}               & ResNet-50 & 7.0  & 21.7  & 2.8    & 1.0  & 4.7  & 11.5 & 13.5 \\
			RetinaNet \cite{ref33}            & ResNet-50 & 8.7  & 22.3  & 4.8    & 2.4  & 8.9  & 12.2 & 16.0 \\
			ATSS \cite{ref19}                  & ResNet-50 & 12.8 & 30.6  & 8.5    & 1.9  & 11.6 & 19.5 & 29.2 \\ \midrule
			RepPoints \cite{ref34}             & ResNet-50 & 9.2  & 23.6  & 5.3    & 2.5  & 9.2  & 12.9 & 14.4 \\
			AutoAssign \cite{ref35}            & ResNet-50 & 12.2 & 32.0  & 6.8    & 3.4  & 13.7 & 16.0 & 19.1 \\
			FCOS \cite{ref36}                  & ResNet-50 & 12.6 & 30.4  & 8.1    & 2.3  & 12.2 & 17.2 & 25.0 \\
			M-CenterNet \cite{ref26}           & DLA-34    & 14.5 & 40.7  & 6.4    & 6.1  & 15.0 & 19.4 & 20.4 \\ \midrule
			Faster R-CNN w/ SimD  & ResNet-50 & 23.9     & 54.5      &  17.6      & 11.9     & 24.8     & 28.0     & 33.5     \\
			Cascade R-CNN w/ SimD & ResNet-50 & 25.0    & 53.2     & 19.1       & 13.2    & 25.8   & 29.0   & 35.7    \\
			DetectoRS w/ SimD     & ResNet-50 & \textbf{26.6}    & \textbf{55.9}    & \textbf{21.2}      &  \textbf{13.4}    & \textbf{27.5}    & \textbf{30.9}    & 37.8  \\ \bottomrule
		\end{tabular}
	\end{table*}
	
	\subsection{Experimental settings}
	
	In the following series of experiments, we use a computer with one NVIDIA RTX A6000 GPU and implement various models based on the object detection framework MMDetection \cite{ref30} and PyTorch \cite{ref31}.
	We use general object detectors such as Faster R-CNN, Cascade R-CNN and DetectoRS as the base models and simply replace the MaxIoUAssigner module with our SimD assignment module.
	Our method can effectively adaptive to any backbone networks and anchor-based detectors.
	Following the main stream settings, for all of the models, ResNet-50-FPN pretrained on ImageNet is used as the backbone, and stochastic gradient descent (SGD) is used as the optimizer, with a momentum of 0.9 and a weight decay of 0.0001. 
	The batch size is set to 2, and the initial learning rate is 0.005.
	The number of RPN proposals is 3000 in both the training and testing stages.
	For the VisDrone2019 dataset, the number of epochs for training is set to 12, and the learning rate decays at the 8th and 11th epochs.
	For AI-TOD, AI-TODv2 and SODA-D, the number of training epochs is 24, and the learning rate decays in the 20th and 23rd epochs.
	For NMS, we use the IoU metric, and the IoU threshold is set to 0.7 for RPN and 0.5 for R-CNN.
	Other aspects of the configuration, such as the data preprocessing and pipeline, follow the defaults in MMDetection.
	
	\begin{table*}[t]
		\centering
		\caption
		{
			Main results on AI-TODv2.
			All of the settings are as same as those for AI-TOD
		}
		\label{AI-TODv2}
		\begin{tabular}{l|c|ccc|cccc}
			\toprule
			Method                   & Backbone       & AP   & AP$_{0.5}$ & AP$_{0.75}$ & AP$_{vt}$ & AP$_{t}$  & AP$_{s}$  & AP$_{m}$  \\ \midrule
			TridentNet \cite{ref32}               & ResNet-50      & 10.1 & 24.5  & 6.7    & 0.1  & 6.3  & 19.8 & 31.9 \\
			Faster R-CNN \cite{ref8}             & ResNet-50-FPN  & 12.8 & 29.9  & 9.4    & 0.0  & 9.2  & 24.6 & 37.0 \\
			Cascade R-CNN \cite{ref24}            & ResNet-50-FPN  & 15.1 & 34.2  & 11.2   & 0.1  & 11.5 & 26.7 & 38.5 \\
			DetectoRS \cite{ref25}                & ResNet-50-FPN  & 16.1 & 35.5  & 12.5   & 0.1  & 12.6 & 28.3 & \textbf{40.0} \\ 
			DetectoRS w/ NWD-RKA \cite{ref23}    & ResNet-50-FPN  & 24.7 & 57.4  & 17.1   & 9.7  & 24.2 & 29.8 & 39.3 \\ \midrule
			YOLOv3 \cite{ref37}                   & DarkNet-53     & 4.1  & 14.6  & 0.9    & 1.1  & 4.8  & 7.7  & 8.0  \\
			RetinaNet \cite{ref33}                & ResNet-50-FPN  & 8.9  & 24.2  & 4.6    & 2.7  & 8.4  & 13.1 & 20.4 \\
			SSD-512 \cite{ref10}                  & VGG-16         & 10.7 & 32.5  & 4.0    & 2.0  & 8.7  & 16.8 & 28.0 \\ \midrule
			RepPoints \cite{ref34}                & ResNet-50-FPN  & 9.3  & 23.6  & 5.4    & 2.8  & 10.0 & 12.3 & 18.9 \\
			FoveaBox \cite{ref38}                 & ResNet-50-FPN  & 11.3 & 28.1  & 7.4    & 1.4  & 8.6  & 17.8 & 32.2 \\
			FCOS \cite{ref36}                     & ResNet-50-FPN  & 12.0 & 30.2  & 7.3    & 2.2  & 11.1 & 16.6 & 26.9 \\
			Grid R-CNN \cite{ref39}               & ResNet-50-FPN  & 14.3 & 31.1  & 11.0   & 0.1  & 11.0 & 25.7 & 36.7 \\ \midrule
			Faster R-CNN w/ SimD     & ResNet-50-FPN  & 24.5     & 56.3      & 17.2      & 10.3  & 24.9  & 28.9    & 35.4     \\
			Cascade R-CNN w/ SimD    & ResNet-50-FPN  & 25.2     & 55.2    & 19.1      & 9.7    & 25.3    & 30.3   & 37.4    \\
			DetectoRS w/ SimD        & ResNet-50-FPN  & \textbf{26.5}    & \textbf{57.7}   & \textbf{20.5}     & \textbf{11.6}    & \textbf{26.7}    & \textbf{30.9}   &  38.4   \\ \bottomrule
		\end{tabular}
	\end{table*}
	
	To facilitate comparison with previous research results, during the testing stage, we use the AI-TOD benchmark
	\begin{table}[t]
		\centering
		\caption
		{
			Main results on VisDrone2019.
			All models are trained on the train set and tested on the val set
		}
		\label{VisDrone2019}
		\begin{tabular}{l|cc|cc}
			\toprule
			Method                   & AP   & AP$_{0.5}$ & AP$_{vt}$ & AP$_{t}$  \\ \midrule
			Faster R-CNN             & 22.3 & 38.0  & 0.1  & 6.2  \\
			Cascade R-CNN            & 22.5 & 38.5  & 0.5  & 6.8  \\
			DetectoRS               & 25.7 & 41.7  & 0.5  & 7.6  \\ \midrule
			DetectoRS w/ NWD-RKA     & 27.4 & 46.2  & 4.4  & 12.6 \\ 
			DetectoRS w/ RFLA        & 27.4 & 45.3  & 4.5  & 12.9 \\ \midrule
			Faster R-CNN w/ SimD     & 26.5    & 47.9     & 7.5  & 16.3    \\
			Cascade R-CNN w/ SimD    & 27.7    & 48.6     & 6.8    & 16.1  \\
			DetectoRS w/ SimD    & \textbf{28.7}   & \textbf{50.3}   & \textbf{7.6}    & \textbf{17.3}    \\ \bottomrule
		\end{tabular}
	\end{table}
	evaluation metrics, which comprise the Average Precision (AP), AP$_{0.5}$, AP$_{0.75}$, AP$_{vt}$, AP$_{t}$, AP$_{s}$ and AP$_{m}$, for AI-TOD, AI-TODv2 and VisDrone2019.
	For the SODA-D dataset, we use the COCO evaluation metrics.
	
	\subsection{Results}
	
	We design four groups of experiments on the AI-TOD, AI-TODv2, VisDrone2019 and SODA-D datasets.
	In each group, we replace the IoU metric with our SimD metric in the RPN module and then apply this module in combination with traditional object detection models, including Faster R-CNN, Cascade R-CNN and DetectoRS.
	
	The results on AI-TOD are shown in Table \ref{AI-TOD}, where we compare our method to several typical object detection methods.
	The detectors in the first seven rows are two-stage anchor-based detectors, those in the next three and four rows are one-stage anchor-based and anchor-free detectors, respectively, and the last three rows show the results of our method.
	Compared with Faster R-CNN, Cascade R-CNN and DetectoRS, we achieve AP improvements of 12.8, 11.2 and 11.8 points, respectively, by using SimD in place of the IoU in RPN.
	We also compare our method with some specialized detectors for tiny objects, namely, DotD, NWD and RFLA, relative to which our method improves the AP by 10.5, 5.8 and 1.8 points, respectively.
	
	The performance of our method on tiny objects is worthy of special attention.
	Because of the extremely small size of these objects (very tiny refers to a size range from 2 to 8 pixels), the AP$_{vt}$ of general object detectors is 0, whereas with the use of SimD, the AP$_{vt}$ values of Faster R-CNN, Cascade R-CNN and DetectoRS are increased from 0 to 11.9, 13.2 and 13.4 points, respectively.
	
\begin{figure*}[t]
		\centering
		\includegraphics[height=6.5cm]{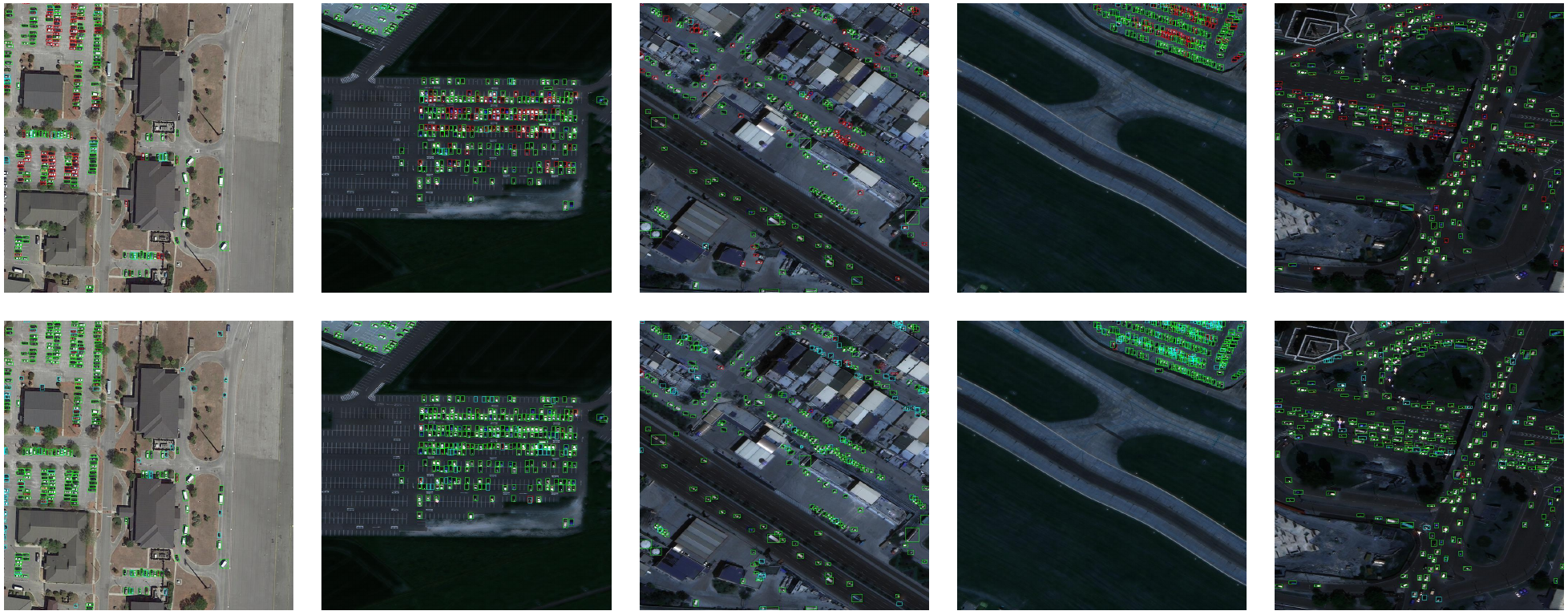}
		\caption{
			Comparison of detection results on AI-TOD dataset between label assignment with the traditional IoU metric and SimD.
			The first row shows the results of Faster R-CNN with the IoU metric, and the second row is also based on Faster R-CNN but with the SimD metric.
			The green, blue and red boxes respectively denote true positive (TP), false positive (FP) and false negative (FN) predictions.
			The improvement achieved with our method is obvious.}
		\label{compare}
	\end{figure*}

	In addition to AI-TOD, our method also achieves the best performance on AI-TODv2, VisDrone2019 and SODA-D, as shown in Table \ref{AI-TODv2}, Table \ref{VisDrone2019} and Table \ref{SODA-D}, respectively.
	On AI-TODv2 and SODA-D, the AP of our method is 1.8 and 1.6 points higher, respectively, than that of its best competitor.
	On VisDrone2019, which contains both tiny and general-sized objects, our method also performs well, in particular, it achieves a 1.3 points improvement over RFLA.
	In Table \ref{SODA-D}, the AP$_{0.5}$ is almost at the same level with RFLA but AP$_{0.75}$ is much higher, this may indicate that our method is more capable of tiny object detection.
	Some typical visual comparisons between the IoU metric and SimD are shown in Fig. \ref{compare}.
	We can find there is an obvious improvement of detection performance after using our method.
	
	\begin{table}[t]
		\centering
		\caption
		{
			Main results on SODA-D.
			All models are trained on the train set and tested on the test set
		}
		\label{SODA-D}
		\begin{tabular}{l|ccc}
			\toprule
			Method                 & AP   & AP$_{0.5}$ & AP$_{0.75}$ \\ \midrule
			Faster R-CNN \cite{ref8}                 & 28.9 & 59.7  & 24.2   \\
			Cascade R-CNN \cite{ref24}                & 31.2 & 59.9  & 27.8   \\
			RetinaNet \cite{ref33}                   & 28.2 & 57.6  & 23.7   \\
			FCOS \cite{ref36}                        & 23.9 & 49.5  & 19.9   \\
			RepPoints \cite{ref34}                    & 28.0 & 55.6  & 24.7   \\
			ATSS \cite{ref19}                         & 26.8 & 55.6  & 22.1   \\
			YOLOX \cite{ref40}                      & 26.7 & 53.4  & 23.0   \\
			RFLA \cite{ref4}                         & 29.7 & \textbf{60.2}  & 25.2   \\ \midrule
			Faster R-CNN w/ SimD        & 31.1    & 59.8     & 28.1      \\
			Cascade R-CNN w/ SimD      &  32.4   & 59.2    &  30.5     \\
			DetectoRS w/ SimD           & \textbf{32.8}     & 59.4     & \textbf{31.3}       \\ \bottomrule
		\end{tabular}
	\end{table}
	
	\begin{table}[t]
		\centering
		\caption
		{
			Ablation study on AI-TOD based on Faster R-CNN
		}
		\label{ablation}
		\begin{tabular}{@{}cc|ccc@{}}
			\toprule
			norm\_width & norm\_height & AP   & AP$_{0.5}$ & AP$_{0.75}$ \\ \midrule
			&              &   20.4   &  49.6     &  13.2      \\
			\checkmark	&              &   22.1   &  52.0     &  14.8      \\
			&    \checkmark          &  22.0    &   52.3    &  14.5     \\
			\checkmark	&    \checkmark          & \textbf{23.9} & \textbf{54.5}  & \textbf{17.6}   \\ \bottomrule
		\end{tabular}
	\end{table}
	
	\subsection{Ablation Study}
	In our proposed method, an important operation is normalization based on the width and height of the ground truth and anchor.
	To verify the effectiveness of the normalization operation, we conduct a set of ablation study.
	As shown in Table \ref{ablation}, we respectively compare not normalizing, normalizing only width, only height, and both width and height. 
	The experimental results show that normalization operation achieves a 3.5 points improvement, mainly benefit by the ability to adapt to objects of different sizes in a dataset, and the normalization parameters $m$, $n$ can be adaptively adjusted according to different datasets.

	\subsection{Analysis}
	
	From the experimental results shown in Table \ref{AI-TOD} to Table \ref{SODA-D}, we find that our method achieves the best AP on all four datasets.
	In addition, on the AI-TOD, AI-TODv2, and VisDrone2019 datasets, our method achieves the best results for very tiny, tiny and small objects.
	Three main achievements of our method can be summarized.
	
	First, our method effectively solves the problem of low accuracy for tiny objects.
	The most fundamental reason is that our method fully accounts for the similarity between two bounding boxes, including both location and shape similarity, therefore, only the highest-quality anchors will be chosen as positive samples when using the SimD metric.
	Compared with VisDrone2019, the performance improvements on AI-TOD and AI-TODv2 are more obvious because the objects are much smaller in these two datasets, this phenomenon may also reflects the effectiveness of our method on tiny object detection.
	
\begin{figure}[t]
		\centering
		\includegraphics[height=5cm,width=8cm]{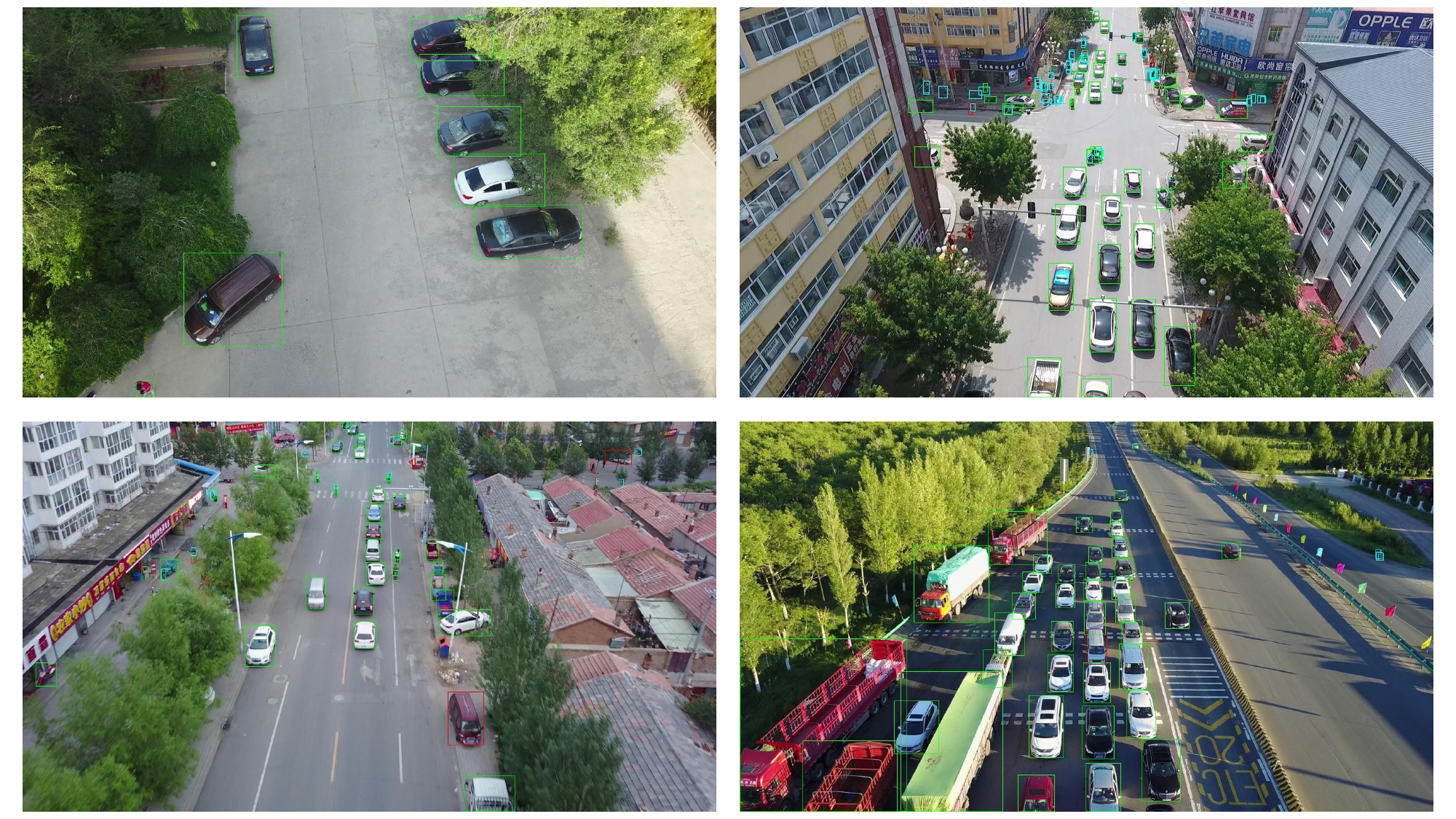}
		\caption{
			Some typical detection results on VisDrone2019 val set, which contains both tiny and general-sized objects.}
		\label{compare_visdrone2019}
	\end{figure}

	Second, our method can adapt well to different object sizes in a dataset.
	In Table \ref{VisDrone2019}, both the AP and AP$_{vt}$ values of our method are the best and much higher than those of other methods.
	The main reason is that normalization is applied in the SimD metric when calculating the similarity between bounding boxes, so it can eliminate the differences arising from bounding boxes of different sizes.
	Some typical detection results are shown in Fig. \ref{compare_visdrone2019}.
	
	Finally, our method achieves the state-of-the-art results on four different datasets.
	Although the characteristics of the objects in different datasets vary, we use the relationships between the ground truths and anchors in the train set when computing the normalization parameters, enabling our metric to automatically adapt to different datasets.
	In addition, there are no hyperparameters that need to be set in our formula.
	
	\section{Conclusion}
	
	In this paper, we point out that most of the existing methods may not be able to automatically adapt to objects of different sizes and include some hyperparameters need to be chosen.
	To this end, we propose a novel evaluation metric named Similarity Distance (SimD), which not only considers both location and shape similarity but also can automatically adapt to different datasets and different object sizes in a dataset.
	In addition, there are no hyperparameters in our formula.
	Finally, we conduct extensive experiments on four classic tiny object detection datasets, in which our method achieves the state-of-the-art results.
	Although our proposed SimD metric is adaptive, it also based on the existing label assignment strategy with fixed thresholds.
	In the future, we aim to further improve the effectiveness of label assignment for tiny object detection.

\end{document}